\pdfoutput=1

\documentclass[11pt]{article}

\usepackage[preprint]{coling}

\usepackage{times}
\usepackage{latexsym}

\usepackage[T1]{fontenc}

\usepackage[utf8]{inputenc}

\usepackage{microtype}

\usepackage{inconsolata}

\usepackage{graphicx}

\usepackage{mathtools}
\usepackage{algorithm}
\usepackage{algorithmic}
\usepackage{listings}
\usepackage{microtype}
\usepackage{multirow}
\usepackage{hyperref}
\usepackage{booktabs}
\usepackage{xspace} 
\usepackage{xcolor}
\usepackage[shortlabels]{enumitem}

\usepackage{xcolor}
\usepackage{subcaption}
\usepackage{pifont}
\usepackage{mystyle}
\usepackage{mlsymbols}
\usepackage{times}
\usepackage{amssymb} 

\definecolor{mdgreen}{rgb}{0.05,0.6,0.05}
\definecolor{darkcyan}{rgb}{0.0, 0.55, 0.55}
\newcommand{\lbl}[1]{\textsc{#1}}
\newcommand{\dataset}[1]{\textcolor{black}{\underline{\textsc{#1}}}}

\newcommand{\TNONE}{\textcolor{darkcyan}{\lbl{None}}}
\newcommand{\TKET}{\textcolor{darkcyan}{\lbl{KpTg}}\xspace} 
\newcommand{\TGSR}{\textcolor{darkcyan}{\lbl{GStuR}}\xspace} 
\newcommand{\TRES}{\textcolor{darkcyan}{\lbl{Restat}}\xspace} 
\newcommand{\TREV}{\textcolor{darkcyan}{\lbl{Revoic}}\xspace} 
\newcommand{\TPRA}{\textcolor{darkcyan}{\lbl{PrsAcc}}\xspace} 
\newcommand{\TPRR}{\textcolor{darkcyan}{\lbl{PrsRea}}\xspace} 

\newcommand{\SNONE}{\textcolor{blue}{\lbl{None}}\xspace}
\newcommand{\SRAS}{\textcolor{blue}{\lbl{RelTo}}\xspace}
\newcommand{\SAMI}{\textcolor{blue}{\lbl{AskMI}}\xspace}
\newcommand{\SMAC}{\textcolor{blue}{\lbl{MClaim}}\xspace}
\newcommand{\SPRE}{\textcolor{blue}{\lbl{PrsEvi}}\xspace}

\newcommand{\TB}{\dataset{TalkMoves}\xspace}
\newcommand{\NCTE}{\dataset{NCTE-119}\xspace}
\newcommand{\SAGA}{\dataset{SAGA22}\xspace}

\title{Enhancing Talk Moves Analysis in Mathematics Tutoring through Classroom Teaching Discourse}

\author{
 \textbf{Jie Cao\textsuperscript{1,2}}\thanks{This work was partially done when Jie Cao was a postdoctoral researcher at the University of Colorado Boulder.},
 \textbf{Abhijit Suresh\textsuperscript{2}},
 \textbf{Jennifer Jacobs\textsuperscript{2}},
 \textbf{Charis Clevenger\textsuperscript{2}},
\\
 \textbf{Amanda Howard\textsuperscript{2}},
 \textbf{Chelsea Brown\textsuperscript{2}},
 \textbf{Brent Milne\textsuperscript{3}},
 \textbf{Tom Fischaber\textsuperscript{3}},
\\
 \textbf{Tamara Sumner\textsuperscript{2}},
 \textbf{James H. Martin\textsuperscript{2}}
\\
\\
 \textsuperscript{1}School of Computer Science, University of Oklahoma,\\
 \textsuperscript{2}Institute of Cognitive Science, University of Colorado Boulder,\\
 \textsuperscript{3}Saga Education
\\
\small{
  \texttt{\{jie.cao\}@ou.edu,}
  \texttt{\{firstname.lastname\}@colorado.edu,}
  \texttt{\{bmilne, tfischaber\}@saga.org}
  }
}

\begin{document}
\maketitle

\begin{abstract}
Human tutoring interventions play a crucial role in supporting student learning, improving academic performance, and promoting personal growth. This paper focuses on analyzing mathematics tutoring discourse using talk moves—a framework of dialogue acts grounded in Accountable Talk theory. However, scaling the collection, annotation, and analysis of extensive tutoring dialogues to develop machine learning models is a challenging and resource-intensive task. To address this, we present \SAGA, a compact dataset, and explore various modeling strategies, including dialogue context, speaker information, pretraining datasets, and further fine-tuning. By leveraging existing datasets and models designed for classroom teaching, our results demonstrate that supplementary pretraining on classroom data enhances model performance in tutoring settings, particularly when incorporating longer context and speaker information. Additionally, we conduct extensive ablation studies to underscore the challenges in talk move modeling.

\end{abstract}


\section{Introduction}
\label{sec:intro}

Human tutoring has become an essential component in combating learning loss due to the COVID-19 pandemic~\cite{robinson2021high,zhou2021cost,engzell2021learning, lewis2021research, patarapichayatham2021covid}. In addition, expanding the tutoring workforce is critical to addressing teacher shortages. However, novice tutors lack adequate training in both their content area and in current pedagogical approaches and thus require extensive professional development. 

Most methods for offering feedback to teachers rely on skilled human observers~\cite{correnti2015improving,wolf2005classroom}, making them costly, time-intensive, and generally inaccessible to paraprofessional tutors. However, recent research has demonstrated automated techniques to reliably detect educationally important discursive features such as productive dialogue, instructional talk, authentic questions, elaborated evaluation, and uptake~\cite{kelly2018automatically,suresh2018using,song2020automatic,demszky2021measuring,jensen2020toward}. 

Much of this earlier work focuses on traditional classroom settings, not small group tutoring. Here, we address the question of whether models initially created for the classroom can serve as the basis for new models for the tutoring setting. This work mainly focus on the creation of discourse analysis tool for mathematics tutoring. Specifically, we focus on {\bf talk moves} --  a set of dialogue acts based on Accountable Talk theory~\cite{o2015scaling, resnick2018accountable, michaels2015conceptualizing}, including both teacher and student talk moves~(see \S\ref{ssec:talkmove-cat}). Research has shown that appropriate use of talk moves in the classroom promotes student learning~\cite{resnick2010well,walshaw2008teacher,webb2019teacher}, and ensure that all students have equal access to participation, subject matter content, and developing appropriate habits of mind~\cite{michaels2008deliberative,o2019supporting}.

To address the mismatch between the classroom and tutoring settings, we developed a new mathematics tutoring dataset with talk move annotations on 121 tutoring sessions. We then examined existing modeling strategies and datasets for classroom mathematics teaching, and explore the best {\em transfer learning strategies} for our target domain. Our modeling experiments and analyses demonstrate how best to use a supervised pretraining-finetuning framework on tutoring talk move analysis, including dialogue context, speaker information, and training strategies. Our best new models outperform existing baselines by a large margin in the tutoring domain and approach the performance of existing models for the classroom domain. Finally, detailed analyses highlight the challenges and point to future work on discourse modeling for mathematics tutoring.

In short, we (1) introduce a new dataset of talk moves annotated math tutoring sessions, (2) describe talk move models for math tutoring with a thorough comparsion with existing models and datasets, (3) highlight the challenges and future work by extensive ablation studies.

\section{Related Work}
\label{sec:intro-background}
Our contributions on new tutoring datasets and transfer learning from existing models build on two lines of research: existing classroom datasets and talk move models for mathematics education.

\subsection{Dialogue Datasets on Mathematics Education}
\label{ssec:classroom-datasets}

Most publicly available  datasets are based on mathematics classroom instruction. The \textbf{TalkMoves}~\cite{suresh2022talkmoves}  and \textbf{NCTE}~\cite{demszky2022ncte} datasets are annotated dialogue corpora collected from real-world classrooms. TalkMoves is derived from three collections of transcripts: Inside Mathematics~\footnote{\url{https://www.insidemathematics.org}};the Third International Mathematics and Science Study~(TIMSS) 1999 video study~\footnote{\url{http://www.timssvideo.com}}; Video Mosaic~\footnote{\url{https://videomosaic.org}}. National Center for Teacher Effectiveness~(NCTE)~\footnote{\url{https://cepr.harvard.edu/ncte}} conducted a systematic collection of recorded mathematics classroom observations, from 2010 to 2013, over 300 classrooms were filmed, resulting in 1,660 lessons for elementary math. Both the TalkMoves and NCTE datasets have been used extensively to create models of classroom discourse. 

Creating a tutoring dataset as the size and quality of earlier TalkMoves and NCTE efforts is time-consuming and expensive. Limited resources are available for authentic, high quality, tutoring sessions.  CIMA~\cite{stasaski-etal-2020-cima}, TSCC~\cite{caines2022teacher} are one-to-one corpora for tutoring for language learning, either  through crowdsourced role-playing or online private chatroom. MathDial~\cite{macina2023mathdial} collect one-to-one dialogue between an expert annotator as teacher and an LLM that simulates the student. Our study addresses this need by providing a small real-world math tutoring dataset annotated in a manner that is consistent with existing resources.\footnote{Please contact the first author for the code and datasets.}


\subsection{Automatic Talk Moves Analysis }
\label{ssec:rw-talkmove}

\citet{suresh2022talkmoves} report on a set of pre-trained transformer-based~\cite{Vaswani2017AttentionIA} models to provide automatic, personalized feedback on the use of this limited set of talk moves. They fine-tuned BERT~\cite{devlin2019bert}, RoBERTa~\cite{liu2019roberta}, Electra~\cite{clark2020electra} in a classification setting~(with one previous utterances as context, and the target utterance,  without speaker information). In later work, ~\citet{suresh2022fine} employed longer contexts by concatenating previous utterances, target utterance, and subsequent utterances into a single input sequence with ordered utterances. A longitudinal pilot study points to the utility value of this tool for teachers, including its positive impact on their discourse practices over time~\cite{jacobs2022promoting,karla2021}. 

 Recently,~\citet{cao2023comparative} extended talk moves modeling to collaborative learning setting, focusing on how the noisy speech in real-world small-group classroom impacts the student talk move modeling. By providing a description and an example utterance for each talk move type, \citet{wang2023can} introduced an instruction-based method to jointly predict the student talk move label with an explanation. Due to consent issues, we re-implement their work as an in-context learning baseline by using Mistral-0.2-instruct-7B model to replace ChatGPT. We leave more LLM studies as future work.

\section{Datasets}
\label{sec:tutoring-data-collection}

\subsection{Talk Move Categories}
\label{ssec:talkmove-cat}
Accountable Talk theory includes a large number of talk move types with varying frequency of use and likelihood of application. For tractability, the existing TalkMoves dataset focuses on \textbf{7 Teacher Talk Moves}, including keeping everyone together~(\TKET), getting student related~(\TGSR), restating~(\TRES), revoicing~(\TREV), press for reasoning~(\TPRR) or accuracy~(\TPRA), none of the above~(\TNONE) and \textbf{5 Student Talk Moves}, such as making claims~(\SMAC), providing evidence and reasoning~(\SPRE), reacting to others ideas~(\SRAS), asking for more information~(\SAMI), and none of the above~(\SNONE). In this paper, we focus on the above talk moves for data annotation and discourse analysis.

\subsection{Data Collection on Math Tutoring}
\textbf{Saga Education} is a non-profit organization that has forged partnerships with school districts across the U.S. with significant low-income and historically marginalized communities. Saga's tutoring model operates on a hybrid framework, wherein students physically attend sessions within a traditional school classroom. Tutors work remotely, leveraging technology to engage with students effectively. Both tutors and students are equipped with individual computers, facilitating interaction through a virtual workspace. This shared environment integrates video conferencing capabilities with speech, chat messages, digital whiteboards, and other essential tools. These features enable detailed mathematical representations, including charts, graphs, tables, and equations.

\paragraph{\SAGA Dataset} 
Our study is based on a high school dataset collected in 2022 and provided to us by Saga (denoted as \SAGA, using the year to distinguish with future version of data collections).~\footnote{We denote this underlined and uppercase text format~(e.g.,\SAGA) to indicate a talk move dataset.} Institutional Research Boards approved all data collection procedures, and data were only collected from students who provided both personal assent and parent’s  consent. From this dataset, we selected 148 videos for analysis. The videos were manually transcribed and three annotators annotated the transcribed conversations with talk move labels with annotation guidelines adapted for tutoring sessions. On a subset of 10 videos, our inter-annotator agreement on all labels reaches more than 80 Cohen's kappa on most of the talk move labels, with for a slightly lower score of 75 on one of the labels. Within the 148 transcribed videos, we annotated 121 sessions, resulting in 69.7 hours of videos with 33695 teacher utterances and 11115 student utterances with talk moves labels.

\subsection{Talk Move Datasets for Teaching}
In addition to the \SAGA dataset,  we reuse two previously published classroom teaching datasets: the \TB and \NCTE datasets described earlier in the related work section.

\paragraph{\TB} The original \TB dataset~\cite{suresh2022talkmoves} contains 567 mathematics classroom sessions covering a broad array of topics from elementary school to high school. All transcripts in the dataset are human-annotated for 7 teacher and 5 student talk moves. Because the previous work didn't release a validation set, we keep the same 63 sessions in the original test set, and re-split the original training set into 441 training, 63 validation, thus denoting the resulting dataset as \TB.

\paragraph{\NCTE} The original NCTE dataset~\cite{demszky2022ncte} has 1660 sessions in total, however, without any talk move annotation on that. We randomly selected 119 sessions to annotate with talk move labels~(thus denoting as \NCTE with the total number of 119 to distinguish with future annotation releases), which are mainly for elementary school math.

Table~\ref{tbl:overall-statistics} summarizes the overall statistics for the three datasets including the total sessions~(sess), Total number of teacher or tutor utterances~(T-utt), the total number of student utterances~(S-utt), the domain~(E-Teach means Elementary Teaching, H-Tutor means High School Tutotring, while Mix-Teaching means math classroom teaching from elementary school to high schools), the average student number~(S-num) and average session length in minutes~(mins) for all the sessions. 

\begin{table}[t]
\centering
 \setlength{\tabcolsep}{2pt}
 \scriptsize
\begin{tabular}{c|cccc|cc}
\hline
&\multicolumn{4}{c|}{\textbf{Overall}} & \multicolumn{2}{c}{\textbf{Per Session}} \\ \cline{2-7}
&sess & T-utt & S-utt & domain & S-num & len \\ \hline
\TB & 567 & 174168 & 59823 & Mix-Teach & ~20 & 30-55 \\
\NCTE & 119 & 27523 & 7241 & E-Teach &  ~20 & ~50 \\
\SAGA & 121 & 33695 & 11115 & H-Tutor &  2-5 & ~35 \\
\hline
\end{tabular}
\caption{\label{tbl:overall-statistics} 
Datasets Summerization}
\end{table}

\begin{figure}[!ht]
    \centering
    \begin{subfigure}[t]{0.42\textwidth}
         \centering
         \includegraphics[trim=12pt 12pt 12pt 10pt,clip=true,width=\textwidth]{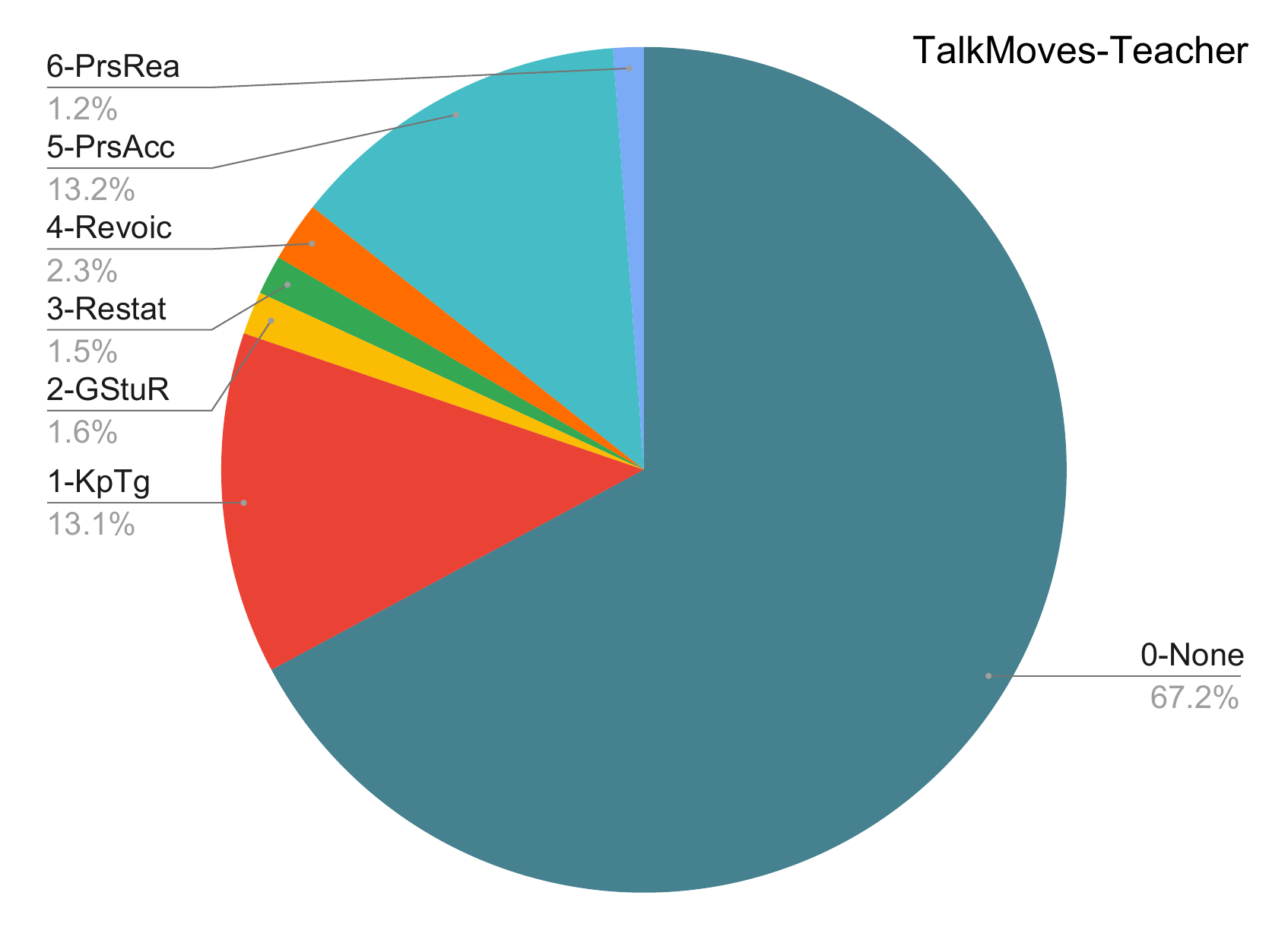}
         \caption{Teacher Talk Moves in \TB}
         \label{fig:teacher-tm-tb}
     \end{subfigure}
         \begin{subfigure}[t]{0.42\textwidth}
         \centering
         \includegraphics[trim=12pt 12pt 12pt 10pt,clip=true,width=\textwidth]{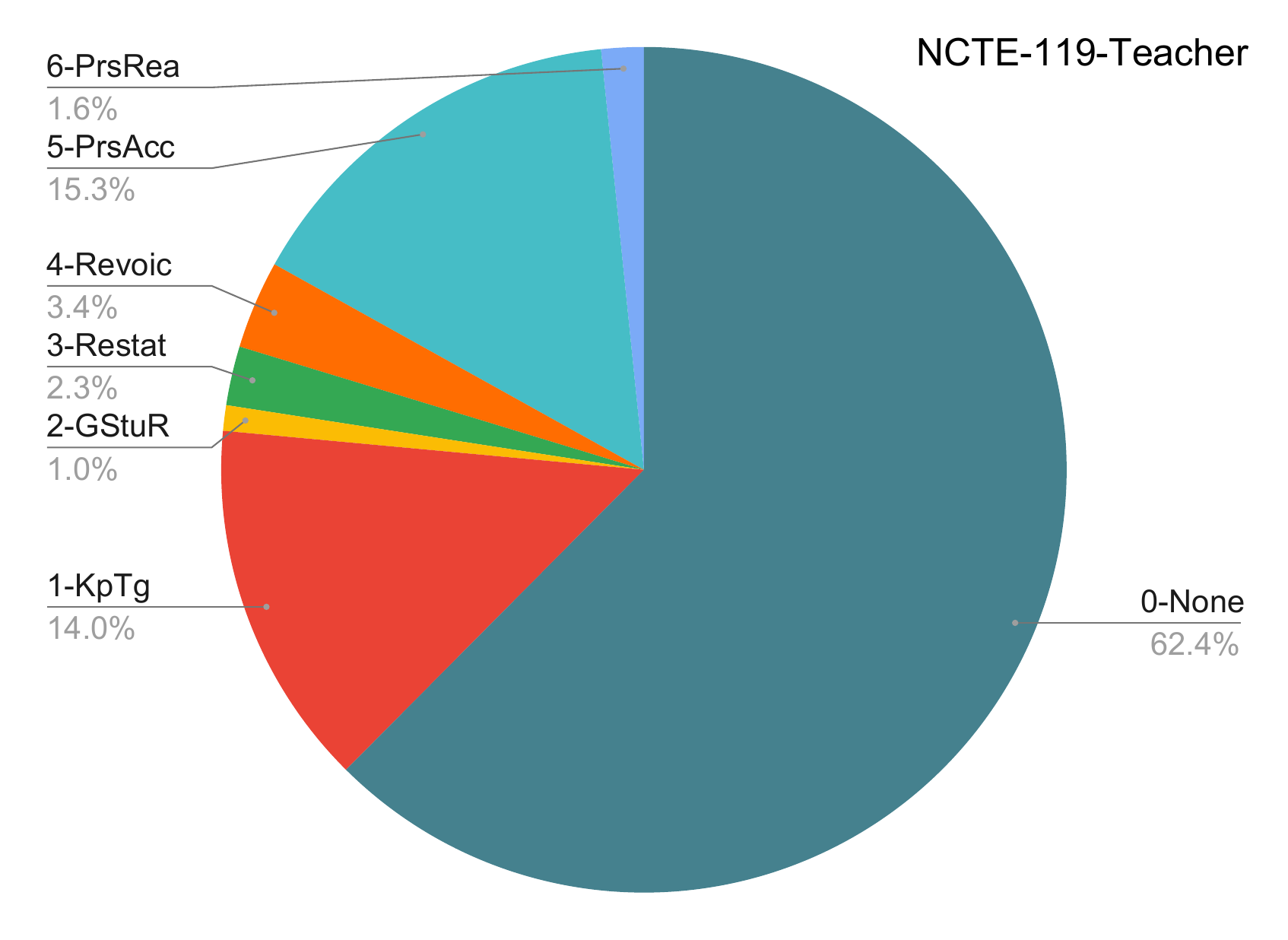}
         \caption{Teacher Talk Moves in \NCTE}
         \label{fig:teacher-tm-ncte}
     \end{subfigure}
    \begin{subfigure}[t]{0.42\textwidth}
         \centering
         \includegraphics[trim=12pt 12pt 12pt 10pt,clip=true,width=\textwidth]{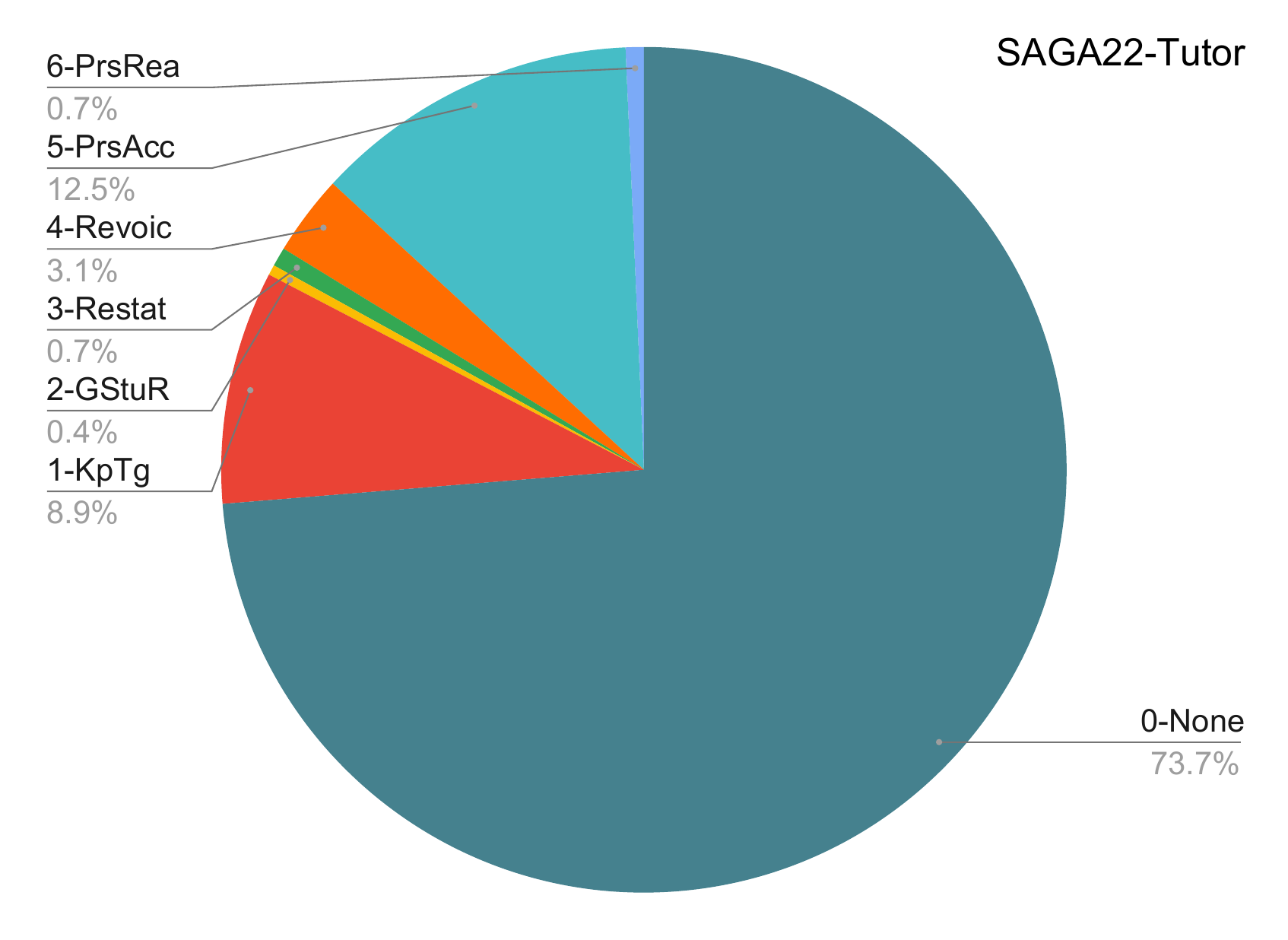}
         \caption{Tutor Talk Moves in \SAGA}
         \label{fig:teacher-tm-saga}
     \end{subfigure}

     \caption{\label{fig:tutor-talk-move-results} Comparison on Teacher/Tutor Talk Moves}
\end{figure}

\begin{figure}[!ht]
    \centering
    \begin{subfigure}[t]{0.42\textwidth}
         \centering
         \includegraphics[trim=12pt 12pt 12pt 10pt,clip=true,width=\textwidth]{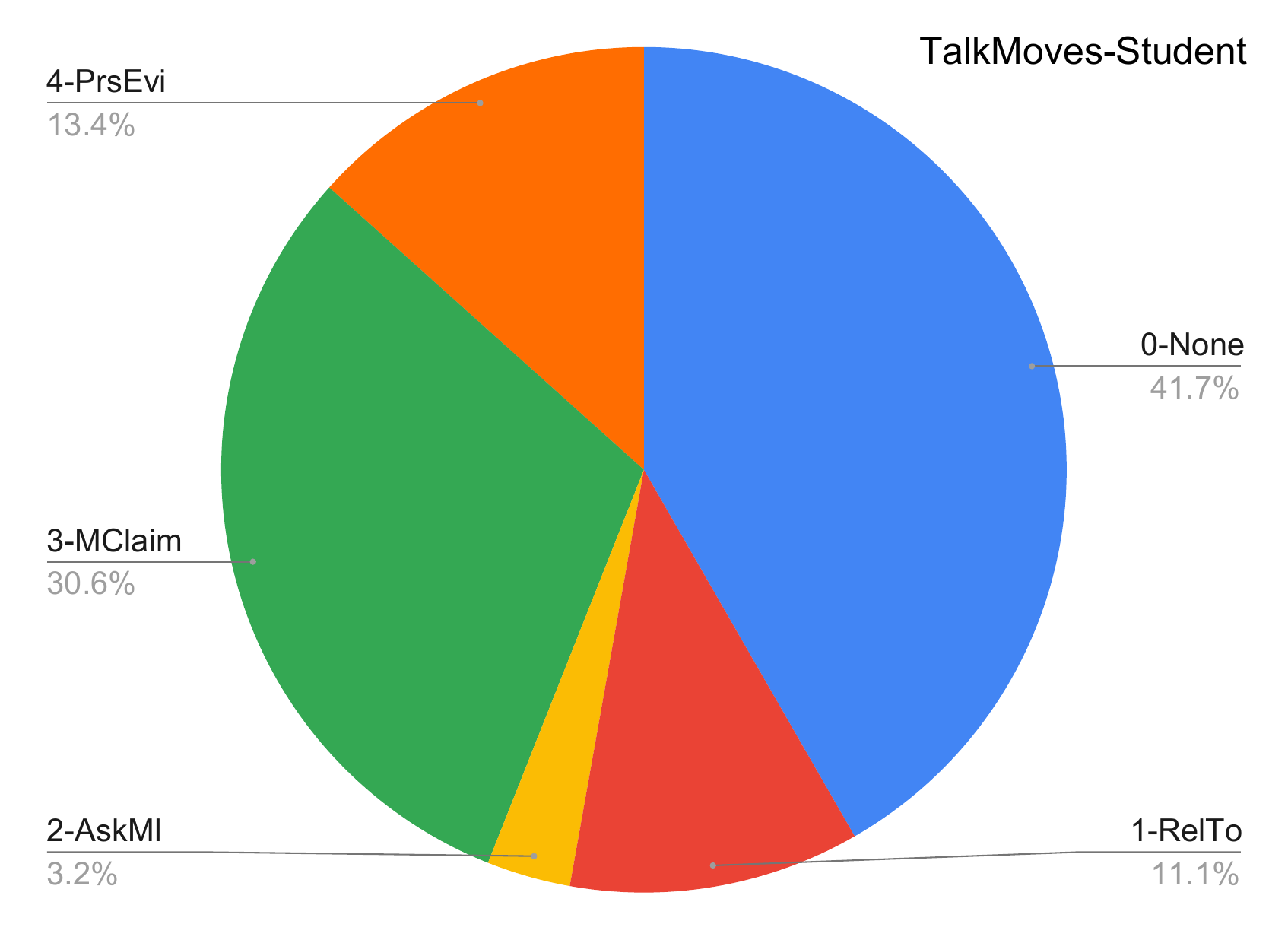}
         \caption{Student Talk Moves in \TB}
         \label{fig:student-tm-tb}
     \end{subfigure}
    \begin{subfigure}[t]{0.42\textwidth}
         \centering
         \includegraphics[trim=12pt 12pt 12pt 10pt,clip=true,width=\textwidth]{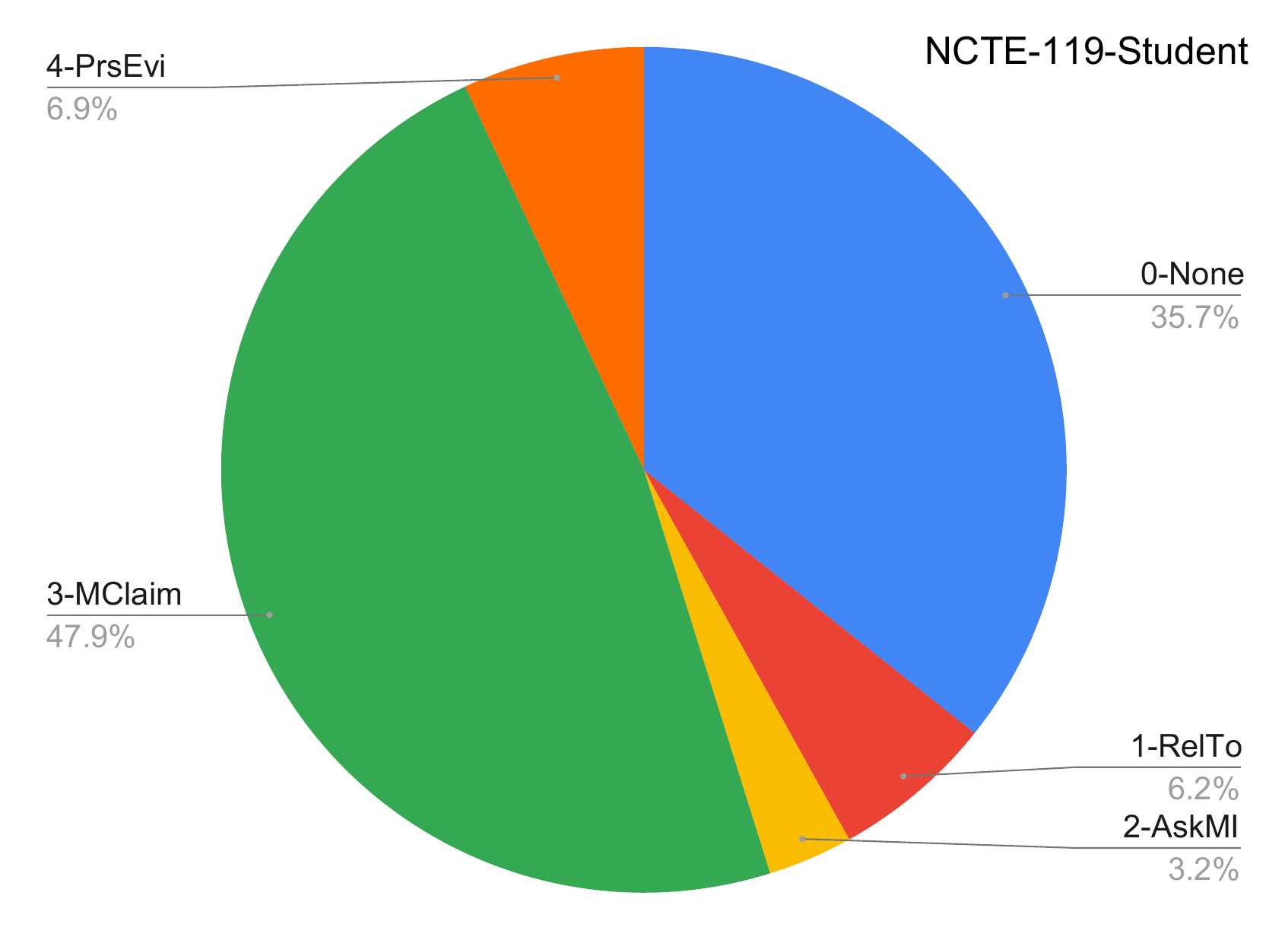}
         \caption{Student Talk Moves in \NCTE}
         \label{fig:student-tm-ncte}
     \end{subfigure}
         \begin{subfigure}[t]{0.42\textwidth}
         \centering
         \includegraphics[trim=12pt 12pt 12pt 10pt,clip=true,width=\textwidth]{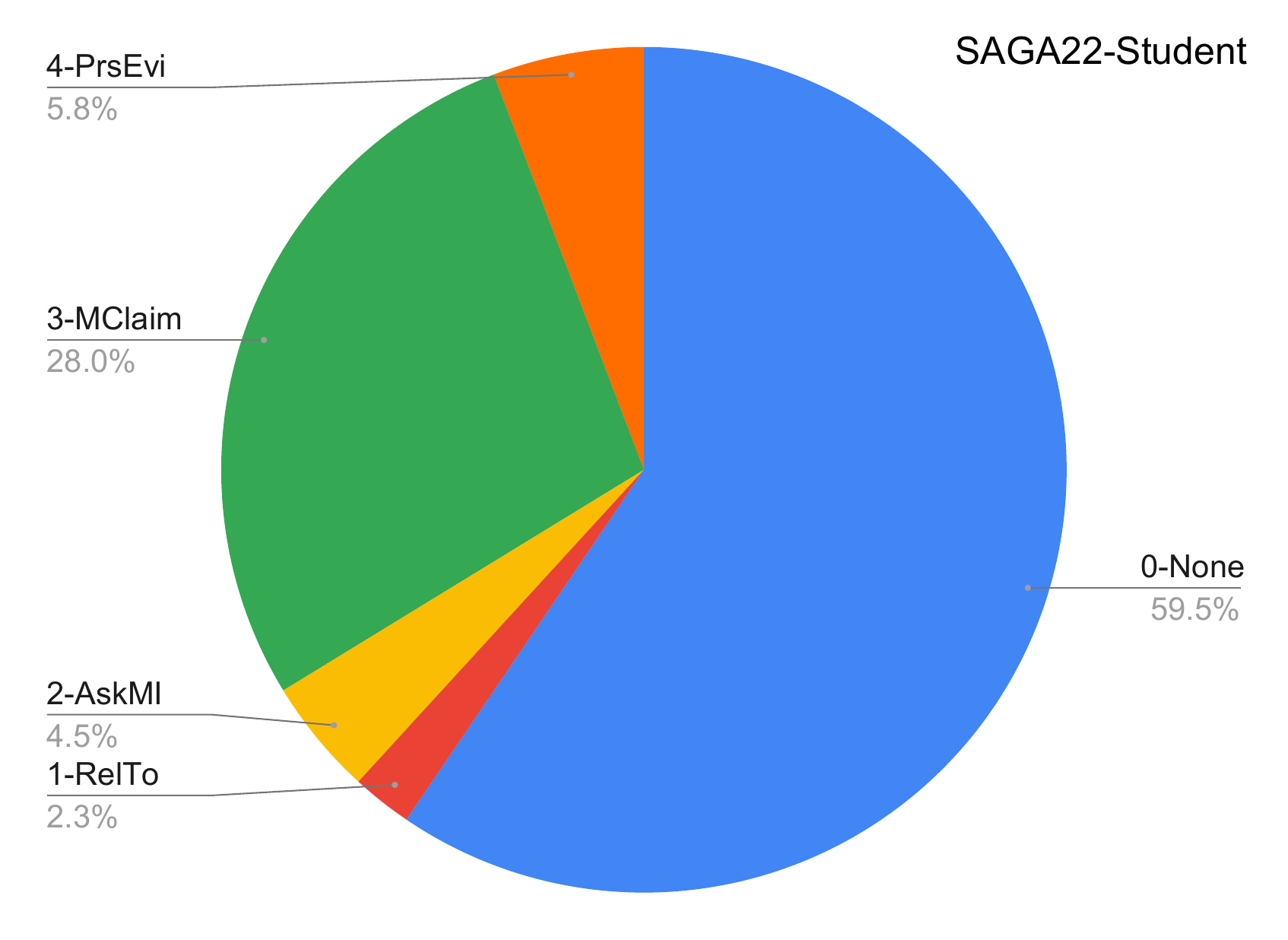}
         \caption{Student Talk Moves in \SAGA}
         \label{fig:student-tm-saga}
     \end{subfigure}
     \caption{\label{fig:stu-talk-move-results}Comparison on Student Talk Moves}
\end{figure}

\subsection{Teaching vs. Tutoring}
\label{ssec:talkmove-compare}
\autoref{fig:tutor-talk-move-results} indicates that tutor talk moves in the \SAGA datasets have the similar distribution with teacher talk moves in \TB and \NCTE. However, \SAGA tutoring setting are  slightly more in~\TNONE~labels, and less in every other talk moves. One possible explanation is that teachers in classroom teaching~(\TB and \NCTE) might receive more training and engage in more proactive pedagogical practices, Alternatively, the grade distribution, with a higher proportion of high school recordings, could result in reduced communication levels~\cite{muhonen2024investigating}. In \autoref{fig:stu-talk-move-results}, students talk moves in tutoring setting are also less than the classroom teaching, which could be indirectly influenced by the reduced use of talk moves by tutors. However, more~\SAMI indicates that small group tutoring provides closer interactions, allowing for more opportunities to ask questions. Overall, teaching and tutoring share similar talk move distribution while differ in various amount.In this paper, we primarily focus on transfer learning, leaving a more in-depth analysis for future work. This will involve exploring latent factors such as class information, grade level, tutor/teacher background, and additional dialogue and discourse analyses~\cite{jurafsky1997switchboard,mann1988rhetorical,asher2003logics,jon-dda2022}.


\section{Models}
\label{sec:models}
Existing models on talk moves analysis could be categorized into two paradigms: pretrain-finetuning~\cite[e.g.,][]{suresh2018using,suresh2022fine}, and in context learning~\cite[e.g,][]{wang2023can}.
We focus on pretrain-finetuning paradigm with "RoBERTa-base" model~\cite{liu2019roberta} as our backbone foundational model~\footnote{Please refer to the appendix~\autoref{sec:roberta-large} for more results and analysis on "RoBERTa-large" models.}. In this paper, our final target task is talk moves analysis for tutoring. Thus, we can use either the original foundational model or the intermediate talk move models designed for teaching as bases for further fine-tuning. When using raw foundational models as bases for target finetuning, we denote it as \textbf{regular pretraining} or \textbf{pretraining from-scratch}. When using intermediate models as bases for target finetuning, we denote the secondary pretraining to build the intermediate models as \textbf{supplementary pretraining}. 

We define a unified model search space $X^{\{\textbf{P},\textbf{F}\}}_{\{\textbf{C},\textbf{SI}\}}$ to cover both teacher/tutor~(when $X=T$), and student~(when $X=S$) models~\footnote{Following previous work, we only consider two separate models for tutor and students respectively, leaving the joint model as future work.}, and their potential improvements. We use \textbf{subscripts} to represent the fundamental modeling settings \textbf{that are consistent} when transiting from pretraining to finetuning, such as (1) dialogue context~(variable~\textbf{C}), (2) speaker information~(variable~\textbf{SI}). We use \textbf{superscripts} to represent training strategies such as (3)~the combinations of supplementary pretraining datasets~(variable ~\textbf{P}) (4) whether further finetuning on \SAGA~(variable~\textbf{F}). Hence, assigning values to all or a subset of 5 variables in $X^{\{\textbf{P},\textbf{F}\}}_{\{\textbf{C},\textbf{SI}\}}$ will lead to a single specific model~($X^{\{\cdot,\cdot\}}_{\{\cdot,\cdot\}}$, with \textbf{a dot $\cdot$} to represent a specific value) or a set of models~($X^{\{,1\}}_{\{\pm7,\}}$, with \textbf{an empty variable value} to represent all possible values for that variable).

\subsection{Dialogue Context: $\textbf{C} \in \{-1,\pm7\}$}
\label{ssec:dial-context}
We follow two settings used in previous works:
\begin{enumerate}
\item \textbf{Previous-One-Utterance}, is used in \citet{suresh2018using,suresh2022talkmoves}, denoted as $-1$. More specific, for teacher models, they use the previous student utterance as context; While, for student models, they use the previous utterance no matter it is from teacher or student. We follow the same sentence pair modeling as the original papers for this context setting, where the context as sequence 1, and the current utterance as sequence 2.
\item \textbf{Previous 7 and Subsequent 7}, is used in \cite{suresh2022fine}, denoted as $\pm7$. We concatenate previous 7 utterances, current utterance, and the subsequent 7 utterances into a single sequence and wrap each utterance as special sentence boundary tokens, thus we keep the original order of the dialogue. Then we force to learn the first special token [CLS] as the context-aware utterance representation of our talk move analysis task~\footnote{Empty utterances will be prepended and padded to make sure there are 15 utterances as inputs; 15 utterance is the longest window size, given we fixed roberta-base model. We decide to keep the two typical and extreme settings $-1$ and $\pm7$ to show the overall trend across a broad range of options.}. 
\end{enumerate}

\subsection{Speaker Information: $\textbf{SI} \in \{spk, nospk\}$}
\label{ssec:speaker-info}
In previous models~\cite{suresh2022talkmoves,suresh2022fine}, the dialogue context didn't use any speaker information during the talk move modeling, which is problematic. For example, without speaker info, when the target utterance simply restating the previous utterance, this could be hard to decide whether it is "Relating to another student" or simply follow the teacher's talk for more information. This could be even worse for longer context settings. Hence, in this paper, we prepend a prefix "T: " or "S: " in front of each utterance to indicate it is said by a teacher/tutor or a student, respectively.
\begin{enumerate}
\item \textbf{Teacher/Tutor Prefix "T:"} is used for both teacher and tutor utterances to make the model easier transferable for the encodings of "T:" from teaching datasets to tutoring datasets. 
\item \textbf{Student Prefix "S:"} is also applied to each students' utterances without distinguishing which student that is. We could only do this for~\TB and~\NCTE datasets, because 20 students are \textbf{not distinguishable} in the classroom session, the transcripts always de-identify the different student speakers as the same student "S: ". Noticing this deficit, we make sure that our~\SAGA transcripts \textbf{explicitly distinguish} different students as "Student-1", "Student-2". However, to be consistent with previous setting, we still use the single student prefix "S:" to model our tutoring dialogue without distinguishing. We leave the transfer learning from bi-party to multi-party as future work.
\end{enumerate}
 When naming the models, we use "spk" and "nospk" as a subscript to indicate with or without speaker prefixes, e.g., $S^{\{,\}}_{\{,spk\}}$ means a set of student models trained with speaker information.

\subsection{Supplementary Pretraining Datasets: $\textbf{P} \in \{\varnothing, ``t",``t+n",``t+s",``t+n+s"\}$}
\label{ssec:pretrain-data}
We have three available talkmove datasets, \TB and \NCTE for teaching, and \SAGA for tutoring. When describing the model name, we use the lower-cased first letter of each dataset name to indicate the pretraining datasets. Since the ~\TB has the largest amount of data, we always involve "t" in our combinations of pretraining datasets, resulting in 4 non-empty combinations and 1 empty pretraining set $\varnothing$, as $\textbf{P} \in \{\varnothing, ``t",``t+n",``t+s",``t+n+s"\}$. We investigate the best combinations for our supplementary pretraining, denoted as a \textbf{superscript}, e.g., $T^{\{t+n+s,\}}_{\{,\}}$ means a family of tutor models pretrained on the combination of all three datasets.

\subsection{\textbf{Fine-tuning on \SAGA: $\textbf{F} \in \{0,1\}$}}
\label{ssec:finetune-indicate}
After the above supplementary pertaining on the combination of datasets, the resultant models could be  inferred on our \SAGA dataset with or without any further fine-tuning. It is unknown which is better. When describing the model name, we indicate this further fine-tuning as a superscript on the model tag~('T' or 'S)', e.g., $T^{\{,1\}}_{\{,\}}$ is a family of models eventually fine-tuned on~\SAGA.

\paragraph{Model Search} 
With the above 5 variables, we first fixed the variable $\textbf{P}="t"$~(given that large amount \TB will be necessary in high probability), and only searched over the rest 16 models assigned with the other four binary modeling choices~(\textbf{X,F,C,SI}), to prioritize the investigation on more interesting modeling factors, such as the dialogue context~(\textbf{C}) and speaker information~(\textbf{SI})). Then we performed extensive search over all other 4 combinations of pretraining datasets~\textbf{P}. In total, we discovered the best models and conducted the ablation studies by searching over 80 experimental settings.

\begin{table*}[t]
\scriptsize
 \setlength{\tabcolsep}{1pt}
\centering

\begin{tabular}{c|c|cccc|cc|ccccccc}
\hline
  Setting&Model &     Context&Speaker&Sup-Pretrain& Finetune&$F_{1}$& Acc & \TNONE & \TKET & \TGSR & \TRES & \TREV & \TPRA & \TPRR \\ \hline
Majority  &$T_{\text{Majority}}$&    N/A&N/A&N/A& N/A&12.1 & 74 & 85 & 0 & 0 & 0 & 0 & 0 & 0 \\ \hline
\multirow{3}{*}{\citet{suresh2022talkmoves}} &$T^{\{\varnothing,1\}}_{\{-1,nospk\}}$&    $-1$&\xmark& N/A&\cmark& 68.6 & 89.9 & 94.5 & 68.7 & 5.6 & 71.8 & 65.9 & 86.6  & 87.0\\ 
&$T^{\{t,0\}}_{\{-1,nospk\}}$&    $-1$&\xmark&\TB& \xmark&76.4 & 89.1 & 93.7 & 69.0 & 56.7 & 73.7 & 68.0 & 84.7 & 88.9  \\
&$T^{\{t,1\}}_{\{-1,nospk\}}$&    $-1$&\xmark&\TB& \cmark& 77.6 & 91.0 & 95.0 & 72.9 & 51.0 & 72.2 & 70.9 & 87.8 & 93.3  \\ \hline
\multirow{3}{*}{\citet{suresh2022fine}} &$T^{\{\varnothing,1\}}_{\{\pm7,nospk\}}$&    $\pm7$&\xmark&N/A& \cmark&58.6 & 68.2 & 93.8& 62.5 & 0.0 & 47.1 & 40.2 & 82.2 & 86.2 \\
 &$T^{\{t,0\}}_{\{\pm7,nospk\}}$&    $\pm7$&\xmark&\TB& \xmark&75 & 89.7 & 94.1 & 71.3 & 50.9 & 73.7 & 58.2 & 87.8 & 88.9 \\
&$T^{\{t,1\}}_{\{\pm7,nospk\}}$& $\pm7$&\xmark&\TB& \cmark&73.7 & 90.7 & 95.0 & \textbf{73.4} & 43.5 & 70.3 & 55.4 & 87.7 & 90.9 \\ \hline
\citet{wang2023can}&ICL-zero-shot& $\pm7$ &\cmark&N/A& \xmark& 24.5 & 18.5 & 3.1 & 30.4 & 14 & 36.5 & 33.3 & 20.0 & 34 \\ \hline
Best From-Scratch&$T^{\{\varnothing,1\}}_{\{-1,spk\}}$&-1    &\cmark&N/A&\cmark &70.6 & 89.8 & 94.5 & 69.2 & 20.5 & 75.7 & 64.4 & 85.9 & 83.6 \\ \hline
Best of All&$T^{\{t+n,1\}}_{\{\pm7,spk\}}$&$\pm7$    &\cmark&\tiny{\TB+\NCTE}&\cmark &\textbf{82.4} & \textbf{91.4} & \textbf{95.1} & 71.6 & \textbf{75.4} & \textbf{81.1} & \textbf{70.3} & \textbf{89.6} & \textbf{93.3} \\
 \hline
\end{tabular}
\caption{\label{tbl:tutor-model-sum} Best tutor models for each setting on \SAGA test set.}
\end{table*}

\begin{table*}[t]
\scriptsize
 \setlength{\tabcolsep}{2pt}
\centering

\begin{tabular}{c|c|cccc|cc|ccccc}
\hline
 Setting&Model &  Context&   Speaker&Sup-Pretrain&Finetune& $F_{1}$ & Acc & \SNONE & \SRAS & \SAMI & \SMAC  & \SPRE  \\ \hline
Majority &$S_{\text{Majority}}$ &  N/A&   N/A&N/A&N/A&15 & 59.7 & 74.8 & 0 & 0 & 0 & 0 \\ \hline
\multirow{3}{*}{\citet{suresh2022talkmoves}}  &$S^{\{\varnothing,1\}}_{\{-1,nospk\}}$ &  -1&   \xmark&N/A&\cmark&61.5 & 82.5 & 89.2 & 0.0 & 67.1 & 77.5 & 71.5 \\
 &$S^{\{t,0\}}_{\{-1,nospk\}}$ &  -1&   \xmark&\TB&\xmark&66.8 & 82.1 & 90 & 25.9 & 69.1 & 77.5 & 71.4 \\
&$S^{\{t,1\}}_{\{-1,nospk\}}$ &  -1&   \xmark&\TB&\cmark&68.2 & 84.9 & 91.0 & 27.1 & 71.4 & 82.5 & 69.0 \\ \hline
 \multirow{3}{*}{\citet{suresh2022fine}} &$S^{\{\varnothing, 1\}}_{\{\pm7,nospk\}}$ &  $\pm7$&   \xmark&N/A&\cmark&45.0 & 74.1 & 85.3 & 0.0 & 30.8 & 66.0 & 43.1 \\
 &$S^{\{t,0\}}_{\{\pm7,nospk\}}$ &  $\pm7$&   \xmark&\TB&\xmark&69.5 & 84.4 & 91.2 & 30.8 & 68.7 & 81.9 & 74.7 \\
&$S^{\{t,1\}}_{\{\pm7,nospk\}}$ &  $\pm7$&   \xmark&\TB&\cmark&69.6 & 85.6 & 91.9 & 29.1 & 70.9 & 82.7 & 73.5 \\ \hline
\citet{wang2023can}&ICL-zero-shot& $\pm7$ &\cmark&N/A& \xmark&25.9 & 27.3 & 4.2 & 40.9 & 24.4 & 34.2 & 25.6 \\ \hline
Best From-Scratch &$S^{\{\varnothing,1\}}_{\{-1,spk\}}$ &  -1&   \cmark&N/A&\cmark&63.6 & 84.9 & 90.8 & 0.0 & 73.5 & 82.5 & 71.7 \\\hline
Best of All &$S^{\{t+n+s,1\}}_{\{\pm7,spk\}}$ &  $\pm7$&   
\cmark&\tiny{\TB+\NCTE+\SAGA}&\cmark&\textbf{76.5} & \textbf{87.4} \multirow{1}{*}{}& \textbf{92.8} & \textbf{48.1} & \textbf{79.0} & \textbf{84.4} & \textbf{78.1} \\
 \hline
\end{tabular}
\caption{\label{tbl:student-model-sum} Best student models for each setting on \SAGA test set.}
\end{table*}


\begin{figure}[t]
    \centering
    \includegraphics[width=0.45\textwidth]{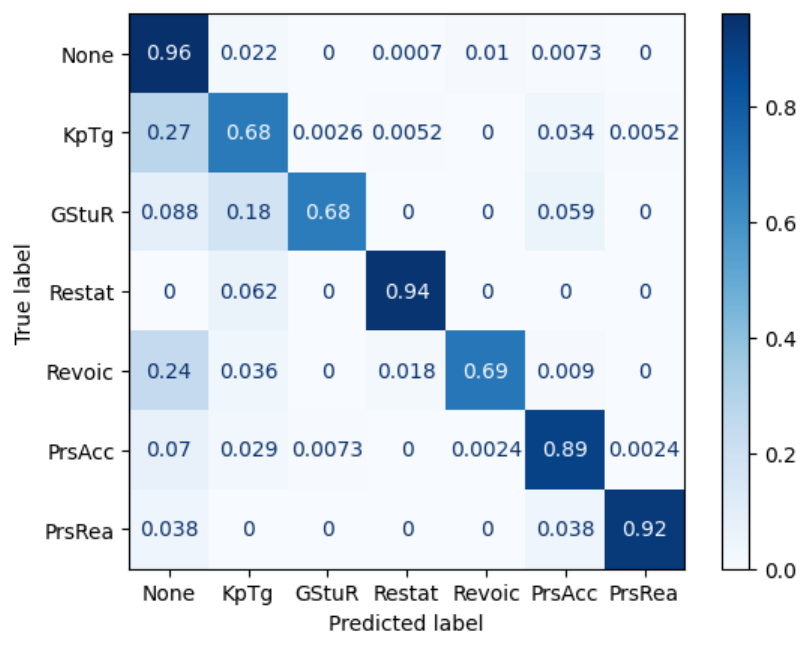}
     \caption{\label{fig:confusion-T} Confusion matrix for the best tutor model $T^{\{t+n,1\}}_{\{\pm7,spk\}}$ on \SAGA}
\end{figure}

\begin{figure}[t]
    \centering
    \includegraphics[width=0.45\textwidth]{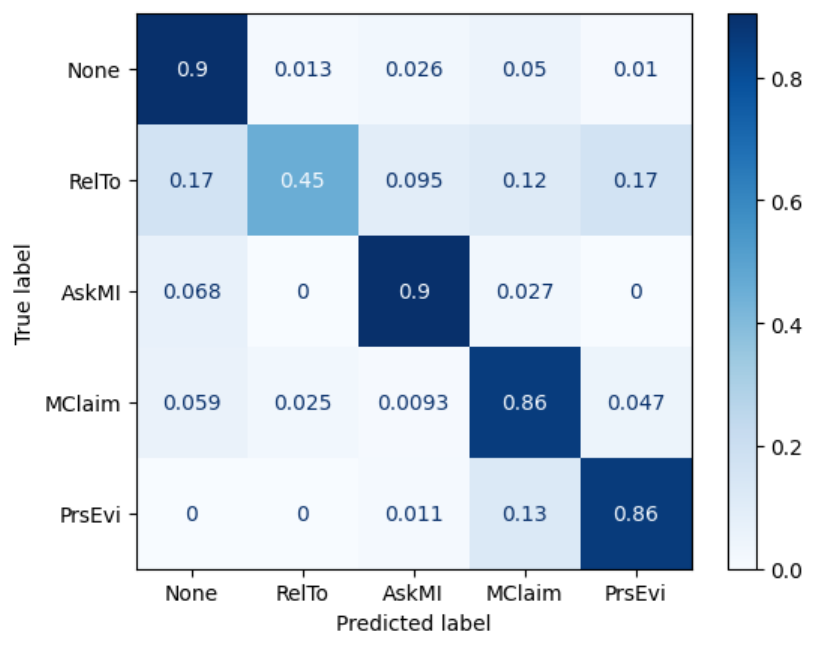}
     \caption{\label{fig:confusion-S} Confusion matrix for the best student model $S^{\{t+n+s,1\}}_{\{\pm7,spk\}}$ on \SAGA.}  
\end{figure}

\section{Results}
\label{sec:results}

With extensive model search, we summarize our main results in Table~\ref{tbl:tutor-model-sum} and~\ref{tbl:student-model-sum} for tutor's and students' talk move analysis respectively. Each table contains 5 categories, the majority baseline, and 2 existing supervised learning work~\cite{suresh2022talkmoves,suresh2022fine}, 1 ICL baseline work~\cite{wang2023can}, and the last 2 rows are the best training from-scratch model and best-of-all model. 
For the two existing supervised learning categories, each work can be extended to three equivalent baseline variants in our model search framework: 
\begin{enumerate}[(1)]
\item \textbf{Trained from-scratch on \SAGA, $X^{\{\phi,1\}}_{\{,\}}$.} We retrain the models from scratch with only \SAGA by following the training strategies described in the corresponding paper. Without any other pretraining, It aims to investigate how the existing modeling strategies perform if training only the 121 tutoring sessions.\vspace{-0.5em}
\item \textbf{Trained on \TB, no finetuning on \SAGA, $X^{\{t,0\}}_{\{,\}}$.} It aims to show that, without finetuning, how the models initially built for math teaching perform on tutoring data.\vspace{-0.5em}
\item \textbf{Pretrained on \TB then finetuning on \SAGA, $X^{\{t,1\}}_{\{,\}}$.} It aims to examine that, with further finetuning on \SAGA, how the models initially built for teaching dataset could be adapted to new tutoring data.
\end{enumerate}
All the best-effort models are selected by the best macro $F_{1}$ score on validation set, and the table here only shows the performance on final evaluation on the held-out test sets.  We show the macro $F_{1}$ score, the accuracy and detailed $F_{1}$ score for each label. Our zero-shot ICL models mimic the similar prompts~(see Appendix \ref{sec:icl}) used in~\cite{wang2023can}, whiling testing on Mistral-0.2-instruct-7B models instead of ChatGPT due to consent issues on our \SAGA datasets. However, they fail to predict the most easiest(frequent) NONE talkmoves in the supervised learning, because it requires complex reasoning to opt out all other labels. This failure causes extremely low macro $F_{1}$ on both tutor and student talkmoves.

\paragraph{Best Tutor Model} Our best tutor model $T^{\{t+n,1\}}_{\{\pm7,spk\}}$ achieves \textbf{82.4 macro $F_{1}$}, reaching the same level of performance of existing models for the classroom domain~\cite{suresh2022fine}. It is firstly pretrained on the combination of teaching-only datasets \TB and \NCTE using previous 7 and subsequent 7 utterances as context, with speaker information, then further finetuned on \SAGA. It achieves the best performance over all talk move categories except for \TKET. With our \SAGA datasets, our best-effort model trained from-scratch can only get 70.6 macros $F_{1}$, particularly failing at predicting \TGSR~(20.5 macro $F_{1}$), which is using a single utterance as context but adding speaker information on that\footnote{As shown in \ref{ssec:dial-context}, adding speaker info to tutor model $T_{-1}$ should be similar with no speaker setting, because the previous sentence must be from the student. Although the prefix "SpeakerName:" is likely to be sparse and not a dominant feature in the training data of RoBERTa, such as OpenWebText~\cite{Gokaslan2019OpenWeb}, but the prefix seems still help.}. Because when we trained models with $\pm7$ context with only the \SAGA dataset, our best-effort model can get a 61.3 macros $F_{1}$ score, and 0 $F_{1}$ score on $\TGSR$, indicating \textbf{the limited \SAGA dataset is not enough to support learning from a longer speaker-aware context}.  \autoref{fig:confusion-T} shows the confusion matrix of best tutor model. The darkness of the diagonal indicates that our model could robustly predict all 7 labels except for \TKET,\TGSR,\TREV. This pattern is highly coorelated with the frequency of each label in our datasets~(see ~\ref{ssec:talkmove-compare}). \textbf{\TGSR are often predicted as \TKET}, because they are relatively similar in the semantics of connecting to students, while the \TGSR is towards a specific student~(on a idea) not all general students. However, our current tutoring model didn't distinguish different student speaker, all students utterances are noted as the same speaker prefix "S:", which calls for better multiparty dialogue modeling.

\paragraph{Best Student Model} Our best student model $S^{\{t+n+s,1\}}_{\{\pm7,spk\}}$ is firstly pretrained on all three datasets using previous 7 and subsequent 7 utterances as context, with speaker information, then further finetuned on \SAGA. It achieves 76.5 macro $F_{1}$, which significantly outperforms all the existing talk move models and best-effort training-from-scratch models, in all student talkmoves, particularly on \SRAS and \SAMI. \autoref{fig:confusion-S} shows the confusion matrix of our best student model. It highlights the failure of predicting \SRAS, which is widely mis-predicted with \SNONE, \SPRE and \SMAC. However, without the help of supplementary training on teaching dataset, our best-effort from-scratch model only achieves 0.0 $F_{1}$ on \SRAS. This poor performance is highly due to its \textbf{rare portion of 2.3\%} as shown in the \autoref{ssec:talkmove-compare}), and the unified student speaker prefix "S:" may also \textbf{lose important discussion thread information between different students}, and causes the model confused with various student behaviors.

 \begin{figure*}[t]
    \centering
       \begin{subfigure}[t]{\textwidth}
         \centering
    \includegraphics[width=\textwidth]{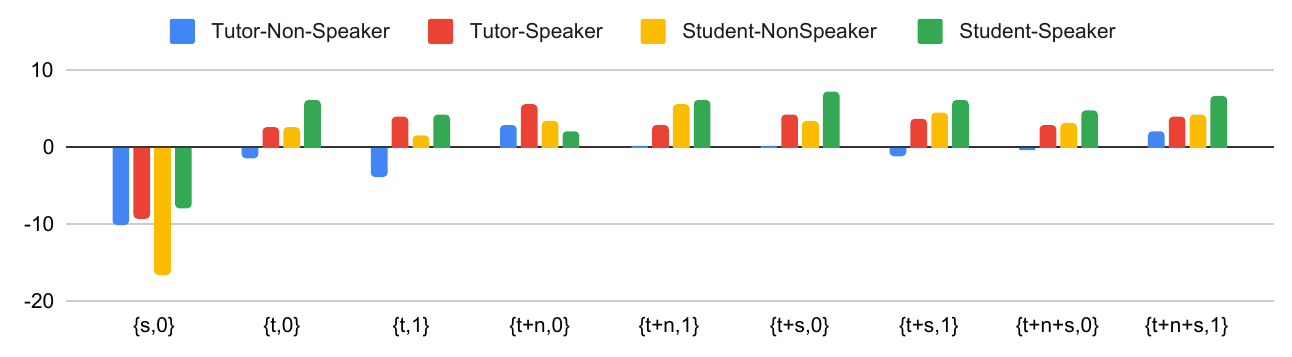}
     \caption{\label{fig:context-ablation}Performance gains when improving the context from $-1$ to $\pm7$.}
     \end{subfigure}

    \begin{subfigure}[t]{\textwidth}
         \centering
        \includegraphics[width=\textwidth]{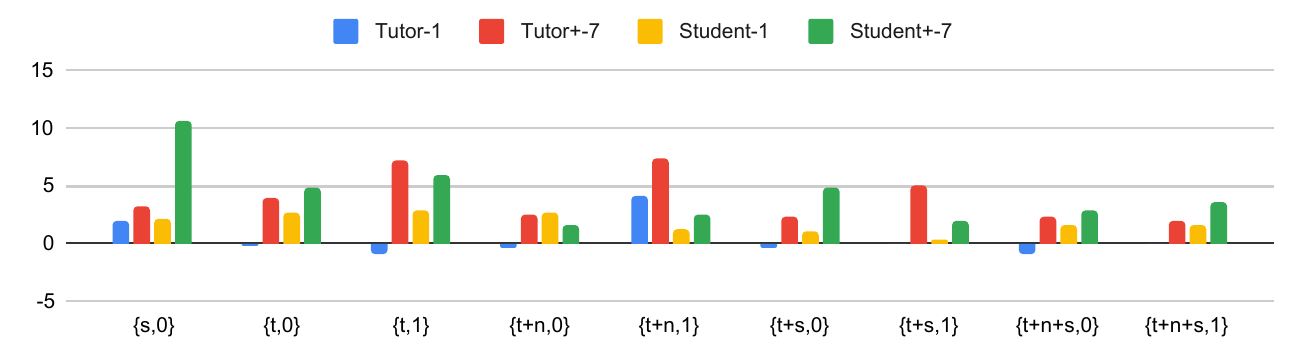}
         \caption{\label{fig:speaker-ablation}Performance gains when adding speaker information.}
    \end{subfigure}
 \caption{\label{fig:speaker-context-ablation}Ablation studies on longer context and speaker information.}    
\end{figure*}

\section{Discussion}
\label{sec:discussion}


We first conduct ablation studies on longer context~(\S\ref{ssec:rst-context}) and speaker info~(\S\ref{ssec:rst-speaker}), respectively.
Then we fixed the best fundamental settings using speaker info with $\pm7$  context, so that we could focus on the impact of supplementary pretraining~(\S\ref{ssec:rst-mix}) and finetuning~(\S\ref{ssec:rst-finetuning}) on the model family of $X^{\{,\}}_{\{\pm7,spk\}}$. 
For five combinations of pretraining datasets~(including $\phi$, without pretraining also can be denoted as $X^{\{s, 0\}}_{\{,\}}$), we order them in increasing size~(s, t, t+n, t+s, t+n+s).

\subsection{Ablation Study on Longer Context} 
\label{ssec:rst-context}
In~\autoref{fig:context-ablation}, each bar value shows the performance change when improving the context from $-1$ to $\pm7$ while keeping the other modeling options consistent by comparing $X^{\{\cdot,\cdot\}}_{\{-1,\cdot\}}$ and $X^{\{\cdot,\cdot\}}_{\{\pm7,\cdot\}}$. Most performance change are positive(ranging from 0.1-7.2), except the \SAGA from-scratch training setting~(the first cluster). Because \SAGA data only is insufficient to support longer context training. Further more, from the left to right, adding the large \textbf{\TB dataset significantly helps to} release the power of long context. However, further adding more \NCTE or \SAGA only have \textbf{diminished returns}. This indicates the longer context in talk move analysis requires sufficient training data to help, but adding more data may not help further. Finally, the performance gains from the model with "Speaker" info~(red and orange) almost always outperform their corresponding "Non-Speaker" variant~(blue and yellow). This indicates that \textbf{simple speaker prefix generally make the longer context more efficient}.

\subsection{Ablation Study on Speaker Information}
\label{ssec:rst-speaker}
We apply a similar method to illustrate the performance changes when adding speaker information to our models. All positive bar values in \autoref{fig:speaker-ablation} shows that \textbf{adding speaker information generally helps all models}, and more significant on $\pm7$ context~(red and orange) than on $-1$ context. More results on teaching-only datasets \TB in~\autoref{sec:extra-talkmoves-results} shows that retraining with longer context and speaker information also outperformed the previous models on \TB in large margin, and RoBERTa-large could further help. However, together with the previous findings in best model analysis, the findings in the longer context ablation study, and the known \textbf{deficit of bi-party speaker prefixes "T:" and "S:"}, we believe fine-grained speaker modeling could support the personalized learning in the tutoring settings better.


\begin{table*}[t]
\scriptsize
\setlength{\tabcolsep}{2pt}
\centering
\begin{tabular}{c|cc|ccccccc|cc|ccccc}
\hline
\multirow{2}{*}{$X^{\{,\}}_{\{\pm7,spk\}}$}& \multicolumn{9}{c}{Tutor Models} & \multicolumn{7}{|c}{Student Models} \\ \cline{2-17}
& $F_{1}$ &Acc & \TNONE & \TKET & \TGSR & \TRES & \TREV & \TPRA & \TPRR & $F_{1}$ & Acc & \SNONE & \SRAS & \SAMI & \SMAC  & \SPRE \\ \hline
$\{s,0\}$ & 61.3 & 88.8 & 94.2 & 64.5 & 0.0 & 47.4 & 51.4 & 85.3 & 86.0 & 55.6 & 80.3 & 89.0 & 0.0 & 53.0 & 76.4 & 79.0 \\
$\{t,0\}$ & 78.9 & 90.1 & 94.4 & 70.8 & 57.1 & 82.1 & 71.3 & 85.9 & 90.7 & 74.4 & 85.5 & 91.7 & 51.4 & 66.7 & 82.2 & 80.0 \\
$\{t,1\}$ & 80.8 & 90.7 & 94.9 & 70.4 & 60.4 & \textbf{88.9} & 70.6 & 87.2 & 93.5 & 75.6 & 87.2 & 92.6 & 44.4 & 75.6 & 84.7 & 80.7 \\
$\{t+n,0\}$ & 81.8 & 90 & 94.2 & 69.5 & 73.8 & 81.1 & 71.7 & 87.6 & \textbf{94.4} & 72.7 & 86.5 & 92.7 & 35.7 & 71.1 & 83.9 & 80.0 \\
$\{t+n,1\}$ & \textbf{82.4} & 91.4 & 95.1 & 71.6 & \textbf{75.4} & 81.1 & 70.3 & \textbf{89.6} & 93.3 & 74.4 & 87.4 & 82.9 & 39.5 & 77.8 & \textbf{84.9} & 76.7 \\
$\{t+s,0\}$ & 80.4 & 91.3 & 95.2 & 73.1 & 63.3 & 80.0 & 73.9 & 88.0 & 89.5 & \textbf{77.1} & \textbf{87.8} & \textbf{93.5} & \textbf{49.5} & 76.3 & 84.8 & \textbf{81.6} \\
$\{t+s,1\}$ & 81.2 & 91.7 & \textbf{95.6} & \textbf{73.7} & 65.6 & 83.3 & 71.8 & 87.3 & 91.3 & 74.5 & 86.6 & 92.6 & 49.4 & 70.9 & 83.2 & 76.2 \\
$\{t+n+s,0\}$ & 80.2 & 91.2 & 95.1 & 72.0 & 66.7 & 75.7 & 73.7 & 88.6 & 89.3 & 76.2 & 87.6 & 93.1 & 46.3 & \textbf{79.7} & 84.8 & 76.9 \\
$\{t+n+s,1\}$ & 81.5 & \textbf{91.7} & \textbf{95.6} & 73.2 & 66.7 & 81.1 & \textbf{74.4} & 87.7 & 91.6 & $76.5^{*}$ & 87.4 & 92.8 & 48.1 & 79.0 & 84.4 & 78.1 \\ \hline
\end{tabular}
\caption{\label{tbl:mixture-finetuning} Model performance for our best fundamental model settings $X^{\{,\}}_{\{\pm7,spk\}}$ on different pretrain-finetuning settings on \SAGA test set. Since the model checkpoints are selected from the validation set, we notice some highlighted numbers are better than our selected best student models in the main results.}
\end{table*}

\subsection{Ablation Study on Pretraining} 
\label{ssec:rst-mix}
Comparing row 2 and row 1 in~\autoref{tbl:mixture-finetuning}, adding \TB into pertaining significantly boosts the performance for every talk move label. Even without using any \SAGA tutoring dataset, the best models in the zero-shot settings~(pretraining without \SAGA and with finetuning tag $\textbf{F}=0$)  perform 81.8 and 74.4 on tutor and student talk moves. It indicates that, with large teaching talk move datasets, \textbf{previous models built for teaching could just work fine on tutoring without finetuning on any the target tutoring data}. Furthermore, adding \NCTE helps on tutor models, generally not on student models~(comparing row 2/3 vs. 4/5, and 6/7 vs. 8/9).
Finally, we noticed that jointly training with \SAGA adds more information about tutoring domain, leading better performance on tutor models, while not on student models. The best performance for each talk move label~(bold numbers) are achieved by different combinations of pretraining datasets, which highlight a  future research direction of \textbf{finding an optimal mixture or data augmentation for lab distributions that could help all labels}.

\subsection{Ablation Study on Finetuning} 
\label{ssec:rst-finetuning}
 Comparing the adjacent rows with the same pretraining datasets~(e.g., $\{t+s, 0\}$ vs. $\{t+s, 1\}$) in~\autoref{tbl:mixture-finetuning},  we noticed that the performance gains for tutor models are \textbf{all positive}~(ranged from 0.6-1.9 $F_{1}$), while not for student models~(finetuning the model pretrained with \TB and \SAGA on~\SAGA again hurt the student model performance). It indicates that \textbf{further fine-tuning may not always help especially when the finetuning dataset is small}. \autoref{sec:roberta-large} shows similar finetuning results with RoBERTa-large , which could further improve tutor models from 82.4 to 85.3, but not on student models. More needs to be done to overcome the \textbf{catastrophic forgetting}~\cite{kirkpatrick2017overcoming} of continuing finetuning as shown in \autoref{tbl:saga-finetuning-forgetting} in \autoref{sec:roberta-large}. Furthermore, \cite{10.1145/3657604.3664664} trained GPT-3.5-turbo to re-write transcripts by appending a label to the end of each tutor utterance, the promising results highlight exciting future directions on LLM-based instruction finetuning.


\section{Conclusion}
\label{sec:conclusion}
In this paper, we investigate how to apply the rich resources on talk move analysis in math teaching to the tutoring domain. We collect a small math tutoring dataset with talk move annotations on 121 tutoring sessions. Then we conduct a thorough examination on existing talk move models on our new tutoring dataset. Based on a unified pretraining-finetuning framework, we systematically search over 4 modeling choices on dialogue context, speaker information, pretraining datasets, and further finetuning to reuse and improve the previous models. We show that without the help on existing models and datasets in the teaching domain, our small amount tutoring data fails to get acceptable performance, and fails to modeling longer context. Our discovered best models with RoBERTa-base achieve 82.4 macro $F_{1}$ on tutor talk moves, and 76.5 on student talk moves. Using RoBERTa-large could further improve the tutor models to 85.3, while not on student models. Lastly, extensive ablations studies show that longer context modeling requires sufficient training data and speaker information support; The current bi-party speaker information always helps; However, better tutoring discourse analysis still calls for future support on modeling multi-party speaker information, optimizing the mixture of pretraining data, and better model finetuning strategies.

\section{Limitations}
\label{sec:limitations}
Our primary focus is on the pretrain-finetuning framework to transfer the model learning from the classroom teaching to math tutoring. 
We keep the same foundational architectures and model components unchanged, such as the speaker information, dialogue context, etc. This is suboptimal for two reasons: (1) to be compatible with existing datasets \TB and \NCTE, where different students are all deidentified as the same student "S". However, our analysis shows that tutoring dialogue have more personalized behaviors and closer interactions, where multiple party dialogue is required. (2) the optimal dialogue context may be also different in the tutoring sessions, and we only demonstrate the premilinary result for ICL methods where "None" label could not be well-classified by 7B public models. We plan to conduct more comprehensive experiments in the future, incorporating longer contexts and LLMs.

Beyond the above modeling limitations, this work is also limited by the datasets themselves. Specifically: (1) the datasets are all from U.S. classrooms with English-only discourse, (2) the domain is limited to mathematics instruction, and (3) the transcripts alone do not provide sufficient context to adequately ground the participants' discourse behavior. Finally, the sensitive nature of the data, including readily available personally identifiable information about teachers, tutors, and students, poses challenges in evaluating potential biases within the models.




\section*{Acknowledgments}
The authors would like to thank the anonymous reviewers for their valuable feedback. This research was supported by the National Science Foundation grant \#2222647 and the NSF National AI Institute for
Student-AI Teaming (iSAT) under grant DRL \#2019805. All opinions are those of the authors and do not reflect those of the funding agencies.


\bibliography{edm,cl,asr,nlp}

\appendix

\section{Results on \TB}
\label{sec:extra-talkmoves-results}
When investigating the two fundamental variables such as the dialogue context~(\textbf{C}) and speaker information~(\textbf{SI})) on talk move analysis, we noticed that the test set of \SAGA is realtively small. To further verify our findings, we also conduct extensive experiments on \TB dataset. \autoref{tbl:tb-results} shows the results on the same test set used in two previous papers on teaching talk moves, where $\{-1,nospk\}$ is equivalent to \citet{suresh2022talkmoves}, while $\{\pm7,nospk\}$ is equivalent to \citet{suresh2022fine}. Combing both speaker information with $\pm7$ dialogue context already outperform the previous models. Furthermore, using RoBERTa-large could achieve the SOTA for both teacher and student models on every talk move label.
\begin{table*}[t]
\scriptsize
\setlength{\tabcolsep}{2pt}
\centering
\begin{tabular}{c|cc|ccccccc|cc|ccccc}
\hline
\multirow{2}{*}{$X^{\{t,0\}}_{\{\cdot,\cdot\}}$}& \multicolumn{9}{c}{Teacher Models} & \multicolumn{7}{|c}{Student Models} \\ \cline{2-17}
& $F_{1}$ &Acc & \TNONE & \TKET & \TGSR & \TRES & \TREV & \TPRA & \TPRR & $F_{1}$ & Acc & \SNONE & \SRAS & \SAMI & \SMAC  & \SPRE \\ \hline
$\{-1,nospk\}$ & 77.2 &87.3	&92.7	&71.5	&61.3	&84.1	&67.7	&82.8	&80.0& 67.8 & 78.1& 84.6	&41.4	&56.2	&78.7	&77.9 \\
$\{-1,spk\}$ & 78.2	&87.8 & 92.8	&72.5	&67.2	&82.9	&67.7	&84.4	&79.6& 71.1 &	80.0	&85.4	&52.2&	57.7	&80.3&	79.7 \\
$\{\pm7,nospk\}$ & 77.0	&88.4	&93.3	&75.6	&64.7	&79.0	&60.1	&85.0	&81.1 &70.1	&80.8	&87.0	&47.0	&54.8	&80.8	&80.7\\
$\{\pm7,spk\}$ & 79.2	&89.0	&93.8	&76.0	&65.3	&83.4	&70.1	&85.4	&80.3 & 73.4&	82.1	&87.9	&58.7	&56.3	&81.8	&82.3 \\ \hline
$\{\pm7,spk\}^{*}$ &  {\bf 81.3}	& {\bf 90.1}	&{\bf 94.5}	&{\bf 78.5}	&{\bf 68.9}	&{\bf 84.9}	&{\bf 73.0}	&{\bf 86.8}	&{\bf 82.4} & {\bf 75.5} &	{\bf 83.9}	&{\bf 89.3}	&{\bf 61.6}	&{\bf 59.8}	&{\bf 83.6}	&{\bf 83.2}\\\hline
\end{tabular}
\caption{\label{tbl:tb-results} Model performance on \TB. * denotes the RoBERTa-large results}
\end{table*}

\begin{table*}[t]
\scriptsize
\setlength{\tabcolsep}{2pt}
\centering
\begin{tabular}{c|cc|ccccccc|cc|ccccc}
\hline
\multirow{2}{*}{$X^{\{,\}}_{\{\pm7,spk\}}$}& \multicolumn{9}{c}{Tutor Models} & \multicolumn{7}{|c}{Student Models} \\ \cline{2-17}
& $F_{1}$ &Acc & \TNONE & \TKET & \TGSR & \TRES & \TREV & \TPRA & \TPRR & $F_{1}$ & Acc & \SNONE & \SRAS & \SAMI & \SMAC  & \SPRE \\ \hline
$\{s,0\}$ & 76.9	&91.8	&96.8&	71.6	&22.2	&92.9&	56.9&	90.5	&93.0& 66.0 &84.8&	93.9&	9.5	&77.9&	81.3&60.2 \\
$\{t,0\}$ & 83.5 & 91.5	&94.6  &77.6	&66.7	&96.4	&65.0	&90.9	&96.5&77.2&	88.2	&90.8	&45.2	&80.2&	88.9&	89.2\\
$\{t,1\}$ & 83.3&92.6&	96.7&	74.9&	52.8&	82.1&	75.6&	89.3&	94.7&75.6	&88.3&	93.3&	33.3&	72.1&	88.4&	83.9\\
$\{t+n,0\}$ & 85.0	&92.2	&95.4&	76.9&	63.9&	100.0&	67.5&	92.5&	94.7&	75.3&	87.7	&91.4&	42.9&	74.4&	88.9&	82.8 \\
$\{t+n,1\}$ & 80.2 &	92.1&	95.7&	75.9&	33.3&	89.3&	73.2&	92.3&	93.0&	77.5&	89.0&	93.2&	35.7&	76.7 &89.6 &	84.9 \\
$\{t+s,0\}$ & 86.1	&93.0&	96.3&	78.1	&63.9&	100.0&	70.7&	91.9&	98.2&	75.7&	88.2&	91.8	&33.3&	82.6&	89.1&	82.8 \\
$\{t+s,1\}$ & \textbf{85.3}	&92.9&	96.1&	76.4&	66.7&	96.4&	74.8&	91.7&	96.5&	\textbf{76.7}&	88.8&	95.1&	35.7&	76.7&	86.3&	79.6  \\
$\{t+n+s,0\}$ & 86.7	&93.1	&97.2&	73.0&	72.2&	96.4&	68.3&	90.7&	98.2&	75.9&	89.1&	93.3&	28.6&	81.4&	89.6&	84.9 \\
$\{t+n+s,1\}$ & 85.4	&92.7	&96.7&	72.3&	63.9&	96.4&	78.9&	90.1&	93.0&	75.7&	88.0&	93.3&	35.7&	82.6&	85.1&	82.8 \\ \hline
\end{tabular}
\caption{\label{tbl:saga-roberta-large} Model performance for our best fundamental model settings $X^{\{,\}}_{\{\pm7,spk\}}$ with different pretrain-finetuning settings on \SAGA test set. All results are based on RoBERTa-large, and comparable to \autoref{tbl:mixture-finetuning}. Since the model checkpoints are selected with the validation set, we notice some numbers are better than the bolded best student models in the main results.}
\end{table*}

\begin{table*}[t]
\scriptsize
\setlength{\tabcolsep}{2pt}
\centering
\begin{tabular}{c|cc|ccccccc|cc|ccccc}
\hline
\multirow{2}{*}{$X^{\{,\}}_{\{\pm7,spk\}}$}& \multicolumn{9}{c}{Tutor Models} & \multicolumn{7}{|c}{Student Models} \\ \cline{2-17}
& $F_{1}$ &Acc & \TNONE & \TKET & \TGSR & \TRES & \TREV & \TPRA & \TPRR & $F_{1}$ & Acc & \SNONE & \SRAS & \SAMI & \SMAC  & \SPRE \\ \hline
$\{t, 1\}$, base & -2.4&	-0.9&	0.8&	-7.6&	-6.8&	-1.9&	1.8&	-2.4&	-0.4& -2.5&	-1.2&	3.3&	-12.2&	-2.8&	-3.3&	-5.6 \\
$\{t, 1\}$, large & -3.5&	-1.4&	1.0&	-9.5&	-1.6&	-17.7&	3.9&	-5.5&	-2.9&-5.6&	-2.5&	6.0	&-22.7&	-33.6&	-9.2&	1.6 \\ \hline
$\{t+n,1\}$,base & -3.4 &	-1.8&	1.7&	-10.5&	-14.4&	-15.0&	-1.3&	-8.4&	-1.1& -0.3 &	-0.3&	5.6	&-5.9&	-7.6&	-6.1&	-1.9\\
$\{t+n,1\}$,large &-1.5	&-1.2&	-0.2&	-3.5&	-13.4&	-11.4&	-1.3&	-2.9&	6.4 & -6.9&	-2.5&	3.9&	-34.0&	-23.3&	-5.7&	3.9\\ \hline
$\{t+s,1\}$,base & -1.9	&-1.4&	-0.6&	-4.9&	-3.9&	-7.8&	0.9&	-0.8&	-8.2 & -2.5 &	-1.1&	3.1&	-13.2&	-7.6&	-4.4&	0.4 \\
$\{t+s,1\}$,large & -1.7&	4.7&	0.3&	-4.3&	-0.7&	-3.9&	2.9&	-2.3&	0.3 & -2.7&	-1.7&	3.2&	-10.9&	-14.4&	-5.8&	-1.9 \\ \hline
$\{t+n+s,1\}$,base & -1.1&	-0.7&	-0.4&	-0.2&	-9.8&	-6.2&	-0.9&	-1.2&	0.7& -1.7	&-0.2&	-1.5&	-7.3&	10.9&	4.0&	-2.2\\ 
$\{t+n+s,1\}$,large & -0.8&	-0.5&	-0.9&	0.7&	1.7&	-3.9&	4.6&	-0.9&	-5.0&-0.4&	-0.8&	1.0&	3.8&	-1.3&	-3.9&	-1.2\\ \hline
\end{tabular}
\caption{\label{tbl:saga-finetuning-forgetting} Model forgetting of the pretrained teaching setting(\TB) after further finetuning the pretrained models on the small \SAGA in the tutor setting.  All models are based our best fundamental model settings $X^{\{,\}}_{\{\pm7,spk\}}$. The numbers show the performance differences of \TB test sets between the models with and without finetuning on \SAGA. "base" and "large" means using RoBERTa-base and RoBERTa-large respectively.}
\end{table*}

\section{Ablation Study on Finetuning with RoBERTa-large}
\label{sec:roberta-large}
Our model search is mainly based on RoBERTa-base. \autoref{tbl:saga-roberta-large} shows the similar ablation studies on finetuning as described in \autoref{ssec:rst-finetuning}, but using RoBERTa-large model. Simply scale-up the model could raise the best model performance on teacher models but not on student models. Furthermore, if comparing the adjacent rows with the same pre-training datasets (e.g., {t + s, 0} vs. {t + s, 1}) in~\autoref{tbl:saga-roberta-large}, we noticed that further finetuning the stronger RoBERTa-large model on our small \SAGA dataset slightly hurts the performance on tutor models, and not stable in student models. 
\autoref{tbl:saga-finetuning-forgetting} shows the performance changes on \TB test set before and after finetuning on the tutoring setting \SAGA, which indicating the forgetting behavior of the pretrained teaching setting(\TB) after further finetuning on the small \SAGA training set in tutoring setting. We noticed the RoBERTa-large model forget more than RoBERTa-base model, which may be related to the relative less training data for the student models, In the future work, we plan to investigate more adaptive finetuning strategies such as LoRA~\cite{hu2021lora} to finetune models on a small dataset settings, and conduct more comprehensive study one this.

\section{In-Context Learning}
\label{sec:icl}
In this paper, we only offer the preliminary studies on prompting methods with longer context and demonstrations. In this section, we describe the detailed prompts we used for each of our setting with running examples to show the possibility of LLMs and highlight the challenges of prompt engineering.

\paragraph{Prompt-based Baseline Models}
As the prompt shown in~\autoref{lst:0-sys-student-prompt},  we first reuse a zero-shot prompt template from~\cite{wang2023can} to predict the talkmoves. The prompt is made by using the label description and examples in Tables 1 and 2 in the original \TB dataset paper~\cite{suresh2022talkmoves}. 

\begin{lstlisting}[basicstyle=\tiny,linewidth=\columnwidth, breaklines=true, label={lst:0-sys-student-prompt}, caption={System Prompt for Student Talk Moves}]
System:
You are a dialogue analyzer to understand the five talk moves for students' utterances, namely "Relating to another student",  "Asking for more information", "Making a claim",  "Providing evidence" and "None". They have the following meaning: "Relating to another student" refers to using commenting on, or asking questions about a classmates' ideas, such as "I did not get the same answer as her."; "Asking for more information" refers to a student requesting more info, saying they are confused or need help, such as "I don't understand number four."; "Making a claim" refers to a student making a math claim, factual statement, or listing a step in their answer, such as "X is the number of cars."; "Providing evidence" refers to a student explaining their thinking, providing evidence, or talking about their reasoning, such as "You can't subtract 7 because they would only get 28 and you need 29."; "None" refers to a student utterance that cannot be classified as one of the four previous talk moves. Considering classifying the  student utterance needs context information, I add its prior student utterances as a context sentence. For example, we need to classify the utterance "Same as you". Predicting this utterance'talk move need its prior sentence "May answer is two" as a context.

So if the prior utterance is "Ah, I thought it was addition". Which kind of talk move the utterance " Me too." belongs to?
\end{lstlisting}

\paragraph{Zero-shot Prompt: Longer Context and Normalized Output}
When conducting prompt-based methods with previous templates, we noticed that different LLMs could have different outputs in quite different formats. To make our analysis easier, we strictly constrain the output format as~\autoref{lst:0-sys-student-prompt-2}, and an example of the input and output example are shown in~\autoref{lst:0-sys-example}. Given the relatively normalized output format, we automatically extract the predicted labels to analyze the generated talk moves and explanations. In this paper, we only use the response explanation to help debug and our prompt engineering. For example, when we use $\pm7$ context for our input example, we found the generated explanation may mistakenly classify other text in the dialogue context instead of the target utterance in the middle.  

\begin{lstlisting}[basicstyle=\tiny,linewidth=\columnwidth,breaklines=true,language=sh, label={lst:0-sys-student-prompt-2}, caption={Noramlizing the output format}]
Considering classifying the target student utterance needs context information, I add its prior and future utterances as dialogue context. Each utterance contains a prefix speaker tag "T:" or "S:" indicating the speaker is a teacher or a student, respectively. Please predict the label of the target text from one of the five talk move labels in the first line as "label: X" and explain the reason in a new line.
\end{lstlisting}

\begin{lstlisting}[basicstyle=\tiny,linewidth=\columnwidth,breaklines=true,stringstyle=\color{mdgreen},label={lst:0-sys-example}, caption={Zero-Shot ICL for Student Talk Move}]
User:
prior_text:
T: What should be our first step
target_text:
S: Subtract A two from both sides
future_text:

Assistant:
label: Making a claim
The target text "Subtract A two from both sides" is a step in solving a math problem, which falls under the "Making a claim" talk move category as the student is stating a step in their solution process.
\end{lstlisting}

Besides that, to make the request more efficient to LLM, we also conduct studies on supporting batch predictions on multiple input examples.

\begin{table}[t]
\scriptsize
\centering
\caption{\label{tbl:iCL-rst} In-Context Learning on \TB}
\begin{tabular}{cccc}
\hline
& & Teacher & Student\\ \hline
Method & Metrics & $F_{1}$ & $F_{1}$\\ \hline
 & Mojority& 12.1& 15.0\\
\multirow{3}{*}{ Zero-ICL } & 0-shot(ChatGPT*)& 37.5* &32.3* \\
& 0-shot(LLama2-7B) & 23.1 &  26.2 \\
& 0-shot(Mixtral-7B)& 24.0 & 25.9 \\ \hline
\multirow{3}{*}{ Few-ICL } & 2-shot(ChatGPT*)& 39.0* & 35.6* \\
& 2-shot(LLama2-7B)& 25.4& 28.4  \\
& 2-shot(Mixtral-7B)& 26.2 & 27.9\\ \hline
\end{tabular}
\end{table}
\subsection{Results on In-Context Learning}
\label{ssec:results-prompts}
Due to the consent issues, we cannot use our \SAGA data in ChatGPT.
Hence, we select a subset of dataset from~\TB to show the baseline performance. Overall, due the limitation of computing resource, we only use 7B version of Llama2 and Mixtral for our ICL testing. As shown in~\autoref{tbl:iCL-rst}, overall, the performance is much worse than the above supervised learning methods. While, the ICL prompt-enginering indeed took more time for manual tuning, instead of the above model search over existing modeling factors. The ChatGPT performance is using the latest gpt-3.5-0125 on a subset of \TB dataset, which is not comparable with our preliminary results on \SAGA test set.




\end{document}